\pdfoutput=1

\documentclass[11pt]{article}
\usepackage{pifont}

\newcommand{\cmark}{\ding{51}}
\newcommand{\xmark}{\ding{55}}

\usepackage[final]{acl}  

\usepackage{times}
\usepackage{latexsym}

\usepackage[T1]{fontenc}

\usepackage[utf8]{inputenc}

\usepackage{microtype}

\usepackage{inconsolata}

\usepackage{graphicx}

\usepackage{amsmath}
\usepackage{multirow}
\usepackage{booktabs}
\usepackage{enumitem}
\usepackage{amssymb}
\usepackage[utf8]{inputenc}
\usepackage{mathrsfs}
\usepackage{mathtools}
\usepackage{microtype}
\usepackage{amsmath,amssymb,amsfonts}
\usepackage{algorithm}
\usepackage{algorithmicx}
\usepackage{algpseudocode}
\usepackage{multirow}
\usepackage{graphicx}
\usepackage{booktabs}
\usepackage{makecell}
\usepackage{colortbl}
\usepackage{xcolor}
\usepackage{enumitem}
\usepackage{graphicx}
\usepackage{amssymb}
\usepackage{pifont}
\usepackage{xcolor}
\usepackage{makecell}
\usepackage{float}
\usepackage{amsmath}
\usepackage{booktabs}
\usepackage{makecell}
\usepackage{latexsym}
\usepackage{times}

\usepackage{microtype}
\usepackage[misc]{ifsym}
\newcommand\freefootnote[1]{%
  \let\thefootnote\relax%
  \footnotetext{#1}%
  \let\thefootnote\svthefootnote%
}
\usepackage[final]{acl}
\usepackage{authblk}

\title{SaSR-Net: Source-Aware Semantic Representation Network for Enhancing Audio-Visual Question Answering}

\author[1]{\textbf{Tianyu Yang}}
\author[2]{\textbf{Yiyang Nan}}
\author[3]{\textbf{Lisen Dai}}
\author[1]{\textbf{Zhenwen Liang}}
\author[4]{\\ \textbf{Yapeng Tian}}
\author[1]{\textbf{Xiangliang Zhang}}
\affil[1]{University of Notre Dame, \texttt{\{tyang4,xzhang33\}@nd.edu}}
\affil[2]{Brown University} 
\affil[3]{Columbia University} 
\affil[4]{The University of Texas at Dallas} 

\begin{document}
\maketitle

\begin{abstract}

    Audio-Visual Question Answering (AVQA) is a challenging task that involves answering questions based on both auditory and visual information in videos.  A significant challenge is interpreting complex multi-modal scenes, which include both visual objects and sound sources, and connecting them to the given question. In this paper, we introduce the Source-aware Semantic Representation Network (SaSR-Net), a novel model designed for AVQA. SaSR-Net utilizes \emph{source-wise learnable tokens} to efficiently capture and align audio-visual elements with the corresponding question. It streamlines the fusion of audio and visual information using spatial and temporal attention mechanisms to identify answers in multi-modal scenes. Extensive experiments on the Music-AVQA and AVQA-Yang datasets show that SaSR-Net outperforms state-of-the-art AVQA methods.
\end{abstract}

\section{Introduction}

\begin{figure}[t]
\centering
\includegraphics[width=0.45\textwidth]{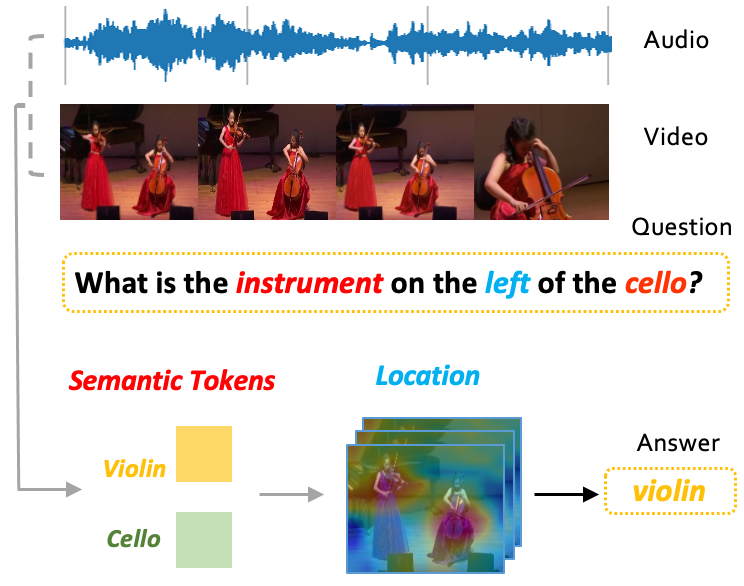}
\caption{Leveraging semantic representation for AVQA involves: (1) Extracting features of various instrument types based on semantic tokens, (2) Identifying the location of the relevant sounding instruments, and (3) Establishing connections between the extracted semantic features, identified instrument locations, and the crucial parts of the question, guiding the model to answer the question accurately.
}
\label{intro}
\end{figure}


Recent contributions to the field of audio-visual question answering (AVQA) include the creation of diverse datasets and sophisticated models \cite{yun2021pano,yang2022avqa,li2022learning,li2023progressive,jiang2023target}. For example, the Pano-AVQA dataset \cite{yun2021pano} contains 360-degree videos paired with corresponding QA sets, while the AVQA-Yang dataset \cite{yang2022avqa} is designed for answering audio-visual questions in real-world scenarios. The MUSIC-AVQA dataset \cite{li2022learning} further broadened the research scope by focusing on spatio-temporal understanding in audio-visual scenes. This dataset uses a dual attention mechanism, identifying sound-producing areas visually first and then applying attention for spatio-temporal reasoning. More recently, PSTP-Net \cite{li2023progressive} was introduced, which progressively identifies key regions relevant to audio-visual questions using refined attention mechanisms.

Existing AVQA methods typically employ general audio and visual encoders to extract features from videos. However, this strategy often fails to link certain sound-producing objects in the video with the responses. Consider questions like \textit{\bfseries What is the instrument on the left of the cello?}
 which necessitates specific type and location awareness, as shown in Fig.~\ref{intro}. Current models often find it difficult to associate the \emph{cello} mentioned in the question with its actual representation in the video scene.

To address these challenges, we propose the Source-aware Semantic Representation Network (SaSR-Net). This model enhances the understanding and integration of individual sound sources and visual objects in AVQA by two strategies: (1) \textbf{Source-wise Learnable Tokens:} Embedded within the Source-aware Semantic Representation Block, these tokens capture essential semantic features from both audio and visual data. This facilitates precise alignment and enhances semantic richness, enabling the model to accurately associate auditory and visual elements based on the query context. (2)  \textbf{Attention Mechanisms:} The model utilizes spatial and temporal attention mechanisms to identify and synchronize relevant visual and audio regions with the query. This not only enhances the accuracy of localization but also strengthens cross-modal associations, crucial for forming a coherent understanding of the scene.




The efficacy of SaSR-Net is demonstrated by its performance on the Music-AVQA ~\cite{li2022learning} and AVQA-Yang ~\cite{yang2022avqa} datasets, where it surpasses state-of-the-art AVQA approaches. The results highlight the effectiveness of the model's source-aware and semantically driven approach in managing complex audio-visual data. Our key contributions are as follows:

\begin{enumerate}
    \item We introduce SaSR-Net, a novel framework that enriches the understanding of sound and visual information, leveraging Source-wise Learnable Tokens to extract semantic-aware audio and visual representations for AVQA.
    \item SaSR-Net integrates multi-modal spatial and temporal attention mechanisms to adaptively leverage visual and audio information in videos for accurate scene understanding.
     \item Our comprehensive experiments and ablation studies demonstrate the effectiveness of our proposed method.
\end{enumerate}

\section{Related Works}

\textbf{Audio-Visual Scene Understanding:} Audio-visual learning focuses on understanding and correlating information from both modalities, aiming to mimic the human's multi-modal perception. This field has been extensively researched in various directions, showing remarkable progress in tasks, \textit{e.g.,} sound source localization 
\cite{hu2021class,liu2022visual,qian2020multiple,mo2023audio}, action recognition \cite{gao2020listen}, event localization \cite{mahmud2023ave,brousmiche2021multi,tian2018audio,zhou2021positive}, video parsing \cite{wu2021exploring,tian2020unified,rachavarapu2023boosting}, captioning \cite{iashin2020multi,tian2019audio}, separation \cite{gao2021visualvoice,tian2021cyclic,zhao2018sound,chen2023iquery}, and dialog \cite{zhu2020describing,alamri2019audio,hori2019end}. Despite this progress, these models still face challenges in integrating the audio modality with visual scene understanding. Effectively leveraging both audio and visual inputs for comprehensive video understanding remains concern. It is essential to consider both audio and visual signals holistically for effective video comprehension. In this work, we propose using Source-wise Learnable Tokens to leverage semantically-aware representations for audio-visual scene understanding.

\noindent
\textbf{Audio-Visual Question Answering:} Audio-Visual Question Answering (AVQA) integrates both modalities, offering a more holistic understanding of scenes. Recent efforts in AVQA include the introduction of datasets such as the Pano-AVQA dataset \cite{yun2021pano}, which features 360-degree videos \cite{yun2021pano}, the real-life AVQA-Yang dataset \cite{yang2022avqa}, and the MUSIC-AVQA dataset \cite{li2022learning}, which focuses on various musical performances \cite{li2022learning}. The MUSIC-AVQA \textit{v2.0} dataset was recently introduced to further reduce dataset bias \cite{liu2024tackling}. Innovations like PSTP-Net \cite{li2023progressive}, which identifies key regions relevant to audio-visual questions through refined attention mechanisms, have been instrumental. Additionally, LAVISH \cite{lin2023vision} introduced a novel parameter-efficient framework for encoding audios and videos, enhancing the potential for practical applications. Despite these advancements, challenges remain in accurately learning video semantics, which can limit the effectiveness of AVQA. Our approach aims to enhance video understanding by modeling semantic entities and strengthening the connections between questions and video content, thereby achieving competitive accuracy.

\section{The Proposed SaSR-Net}
\label{sec:format}

\begin{figure*}[h!]
\centering
\includegraphics[width=1\textwidth]{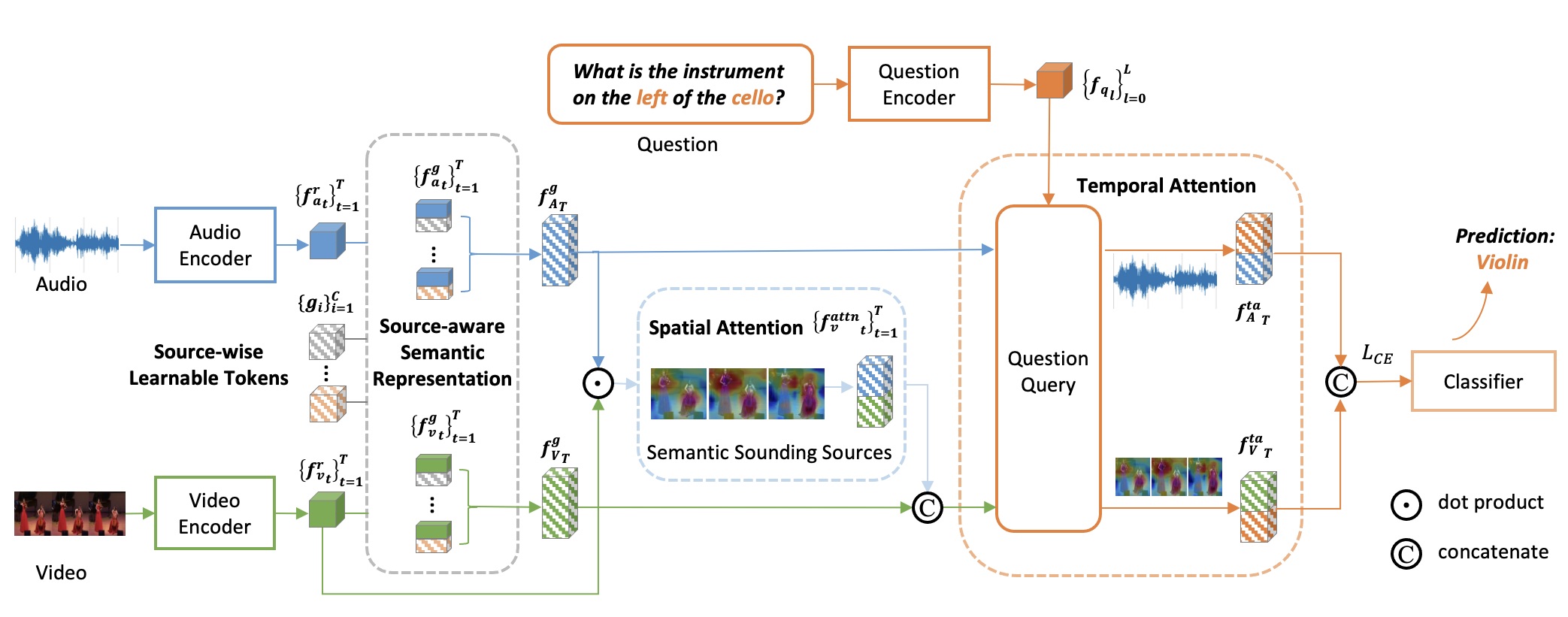}
\caption{The architecture of the proposed SaSR-Net.
} 
\label{fig1}
\end{figure*}

Given a video with both visual and audio tracks, along with a question related to the content within the video, the objective of the AVQA task is to predict an accurate answer response. To achieve this, we propose a novel SaSR-Net architecture. This model is designed to generate compact, semantic-aware embeddings by identifying salient sounding objects present in the audio-visual input that are relevant to the given query. The overview of our proposed framework is illustrated in Figure~\ref{fig1}.

\subsection{Representations for Different Modalities}
\label{sec:representationsfordifferentmodalities}

Given a video with both visual and audio tracks, $V_T$ and $A_T$, we split it into 1-second non-overlapping segment pairs $\{(v_t, a_t)\}_{t=1}^{T}$, where $v_t$ and $a_t$ are the video and audio clips during time $[t - 1, t)$. 
Besides, each sample has a related question $Q_{L} = \{q_l\}_{l = 1}^{L}$ and answer $\mathbf{y}$, \textit{i.e.,} $(\{(v_t, a_t)\}_{t=1}^{T}, \{q_l\}_{l = 1}^{L},\mathbf{y})$, where $q_l$ is a word and  $\mathbf{y}$ is  a one-hot encoding representing  the correct answer.

\noindent
 \textbf{Audio Feature:} Each audio segment \( a_t \) is converted into a raw feature vector \( \mathbf{f}^r_{a_{t}} \) using the pre-trained VGGish \cite{gemmeke2017audio} model, which works on transformed audio spectrograms. In all, the full audio will be transformed to a set of raw feature vectors $\mathbf{f}_{A_T}^r = \{\mathbf{f}^r_{a_t}\}_{t=1}^T$.

\noindent
\textbf{Visual Feature:} Using ResNet-18 \cite{he2016deep}, we process the initial frames from \(V_T\) into raw vectors \(\mathbf{f}_{V_T}^r = \{\mathbf{f}^r_{v_t}\}_{t=1}^T\) and feature maps \(\mathbf{X}^{r}_{PT} = \{{\mathbf{X}^{r}_{P}}_t\}_{t=1}^T 
= \{\{{{\mathbf{x}^{r}_{p}}}_t\}_{p=1}^P\}_{t=1}^T\), where $p$ denotes positions on the feature maps, up to $P$ positions.

\noindent
\textbf{Question Feature:} For a question $Q_L = \{q_l\}_{l = 1}^{L}$, word embeddings are passed through an LSTM.  The resulting feature vectors \( \mathbf{f}_{Q_L} = \{f_{q_l}\}_{l=1}^L \) are derived from the LSTM's final hidden state. Here, $L$ is the max sequence length. The encoder is trained from scratch along with the entire model.

\subsection{Source-wise Learnable Tokens}
\label{sec:tokens}

Distinguishing between audio sources and visual objects in videos fundamentally requires the association of these two modalities. A video may contain several visual objects and sound sources. To accurately respond to questions related to these video scenes, it is essential that our model effectively aligns and associates audio and visual content that are semantically synchronized. To achieve this, we introduce a series of Source-wise Learnable Tokens (SLT). Each token represents a distinct semantic category, such as a \textit{guitar} or \textit{piano}. These tokens will be utilized to align the two modalities and aggregate multimodal source-aware contexts for QA.

We denote Source-wise Learnable Tokens as $\mathbf{G}_C = \{\mathbf{g}_i\}^{C}_{i=1}$. Here, $C$ represents the total number of distinct categories of sounding objects within our dataset.

Initially, we align the Source-wise Learnable Tokens with features from both video and audio by concatenating them.
This computation will help ensure each token matches one of our intended categories, such as guitar or piano.
To achieve this, we prepare category annotations in the labels and guide the model by applying penalties to the tokens during training.
This will be elaborated in the following sections.

Subsequently, we apply self-attention $\mathrm{SelfAttn}$ to aggregate the auditory features $\mathbf{f}^r_{a_t}$ and visual features $\mathbf{f}^r_{v_t}$ separately. Here, we use the notation $[\mathbf{a}; \mathbf{b}]$ to represent the concatenation operation between tensor $\mathbf{a}$ and tensor $\mathbf{b}$, or the split operation between tensor $\mathbf{a}$ and tensor $\mathbf{b}$
\begin{align*}
    [\mathbf{f}^{s}_{a_t}; \mathbf{G}^a_{C}] &= \mathrm{SelfAttn}([\mathbf{f}^r_{a_t}; \mathbf{G}_C]), \\
    [\mathbf{f}^{s}_{v_t}; \mathbf{G}^v_{C}] &= \mathrm{SelfAttn}([\mathbf{f}^r_{v_t}; \mathbf{G}_C]).
\end{align*}

After applying self-attention and splitting, we obtain source-aware audio embedding  $ \mathbf{f}^{s}_{a_t} $, source-aware visual embedding  $ \mathbf{f}^{s}_{v_t} $, and tokens $ \mathbf{G}^a_{C} $ and $ \mathbf{G}^v_{C} $.
In detail, if we assume $D$ is the dimension for each single feature embedding above, the self-attention $\mathbf{S}$ can be represented as ($\mathbf{f}$ is an input feature),
\[
    \mathbf{S}(\mathbf{f}) = \sigma(\frac{\mathbf{f}\cdot\mathbf{f}^{\intercal}}{\sqrt{D}})\cdot\mathbf{f},
\]
where $\sigma$ represents Softmax function.

The obtained representation $ \mathbf{f}^{s}_{a_t} $,  $ \mathbf{f}^{s}_{v_t} $,     $ \mathbf{G}^a_{C} $ and $ \mathbf{G}^v_{C} $ will be used next to compute the source-aware semantic representation.


\subsection{Source-aware Semantic Representation}
\label{representation}

In this section, we assign semantic attention more directly and introduce training penalties to ensure that all learnable tokens accurately represent specific semantic categories.
This design aims to improve our model's capability to precisely represent multi-modal scenes in videos and generate source-aware audio and visual semantic embeddings.

We introduce a source-aware semantic representation block. In the previous section, we have already got both semantically enriched audio and visual embeddings which enhanced with token information. 
Instead of treating the embeddings and Source-wise Learnable Tokens within the same modality as a single entity as we did in Sec.~\ref{sec:tokens}, we hope the model to learn specific information fusion / weighting relationships between the Source-wise Learnable Tokens and the embeddings. As a result, as for the audio/video features that are contained in the embedding and we are also interested in, the model will finally enhance them by properly-learned tokens. To achieve it, we will use our Source-aware Semantic Representation Block to perform cross attention from learnable tokens $ \mathbf{G}^a_{C} $ and $ \mathbf{G}^v_{C} $ to the semantically enriched audio and visual embeddings. 

The resulting semantically-enriched audio embedding $\mathbf{f}^{g}_{A_T}$ = $\{ \mathbf{f}^{g}_{a_t} \}_{t=1}^T$ and video embedding $ \mathbf{f}^{g}_{V_T}$ = $\{ \mathbf{f}^{g}_{v_t} \}_{t=1}^T$ are computed as the following equations performing cross-attention:
\begin{align*}
    & \mathbf{G}^{a\prime}_{C} = \mathbf{G}^a_C + \mathrm{FC}((\mathrm{CrossAttn}( \mathbf{G}^a_C, \mathbf{f}^s_{A_T} )), \\ 
    & \mathbf{G}^{v\prime}_{C} = \mathbf{G}^v_C + \mathrm{FC}((\mathrm{CrossAttn}( \mathbf{G}^v_C, \mathbf{f}^s_{V_T} )), \\
    &  \mathbf{f}^{g}_{A_T} = \mathrm{FC}((\mathrm{CrossAttn}( \mathbf{f}^s_{A_T}, \mathbf{G}^{a\prime}_C )), \\ 
    &  \mathbf{f}^{g}_{V_T} = \mathrm{FC}((\mathrm{CrossAttn}( \mathbf{f}^s_{V_T}, \mathbf{G}^{v\prime}_C)),
\end{align*}
where $\mathbf{f}^s_{A_T}$ = $\{ \mathbf{f}^s_{a_t} \}_{t=1}^T$,  $ \mathbf{f}^s_{V_T}$ = $\{ \mathbf{f}^s_{v_t} \}_{t=1}^T $, $\mathbf{G}^{a\prime}_{C}$ and $\mathbf{G}^{v\prime}_{C}$ are source-aware represented tokens, FC represents a fully-connected layer, LN is layer normalization, and the cross-attention works as: 
\[
    \mathrm{CrossAttn}(\mathbf{a}, \mathbf{b}) = \sigma( \frac{\mathrm{FC}(\mathbf{a}) \cdot \mathrm{FC}(\mathbf{b})}{\sqrt{D}} ) \cdot \mathrm{FC}(\mathbf{b}).
\]
The calculation of cross-attention for $\mathrm{CrossAttn}(\mathbf{G}^{a\prime}_C, \mathbf{f}^s_{A_T})$ and $\mathrm{CrossAttn}(\mathbf{G}^{v\prime}_C, \mathbf{f}^s_{V_T})$ follows the equations above. The fully-connected layer FC is used to align the dimensions of features from different latent spaces.

While the entire set of trainable parameters in SaSR-Net is optimized for minimizing the AVQA loss function that we will define later, it is also important to incorporate auxiliary loss functions specifically targeting the Source-wise Learnable Tokens. These additional loss functions are basically utilizing the prior knowledge to force the Source-wise Learnable Tokens to become the centroids in the hidden space. It will highlight the task-specific significance of these tokens, ensuring that they capture the characteristics of sound sources present in the audio and video. At last, they facilitate the extraction of more meaningful, source-aware representations, which are essential for the AVQA task. 

The first auxiliary loss function is the binary cross-entropy (BCE) loss, which focuses on identifying individual sound sources' presence in the input audio and video channel,
\begin{align*}
    \mathcal{L}_{\mathrm{source}} = & \; \mathrm{BCE}(\sigma(\mathrm{FC}(\mathbf{G}^{a\prime}_C)), \mathbf{p}_{C}) + \\ &
    \mathrm{BCE}(\sigma(\mathrm{FC}(\mathbf{G}^{v\prime}_C)), \mathbf{p}_{C}),
\end{align*}
where $\mathbf{p}_{C}$ is the ground truth label for the source class. This label is compared against the predicted labels generated by applying the sigmoid activation function  $\sigma$ to a fully connected layer, operating on  the semantically enriched audio embedding 
  $\mathbf{f}^{g}_{A_T}$ and video embedding $\mathbf{f}^{g}_{V_T}$. 
  
The second auxiliary loss function serves as a regularization term to ensure that each learned  token uniquely represents a distinct type of sound source. Specifically, we aim for each token vector   $\mathbf{g}_i$ to exclusively represent a single type of sound source. To achieve this, we define the loss using cross-entropy (CE) for sound source classification:
\[
    \mathcal{L}_{\mathrm{reg}} = \mathrm{CE}(\mathrm{FC}(\mathbf{g}_i), \{c\}_{c = 1}^{C}).
\]


\begin{figure*}[h!]
\centering
\includegraphics[width=1\textwidth]{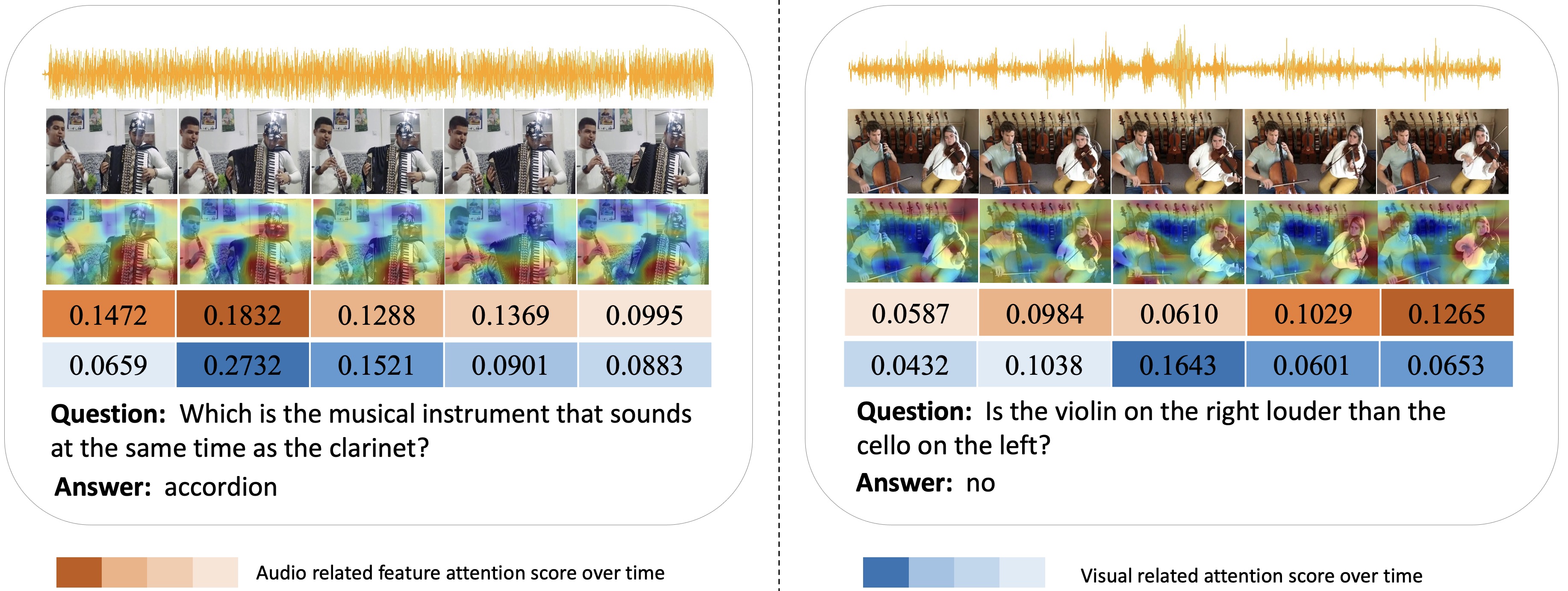}
\caption{Visualization of Spatial Attention (SA) and Temporal Attention (TA) Blocks. The SA Block heatmaps pinpoint sounding object locations, and the TA Block displays audio-visual feature scores. SA localizes critical visual areas, while TA synchronizes video moments with questions, boosting overall audio-visual comprehension.}
\label{sata}
\end{figure*}


\subsection{Multi-modal Spacial Attention}
\label{sec:multimodalspecialattention}

One significant challenge involves localizing visual areas relevant to the given question in the AVQA task. This entails two tasks: firstly, identifying areas with key items by allocating reasonable spatial attention on the visual feature map, and secondly, establishing a temporal connection between the weighted feature map and the question.

Fortunately, the sections from \ref{sec:representationsfordifferentmodalities} to \ref{representation} have already provided us with semantic-aware audio and visual embeddings. The semantic information in these embeddings proves beneficial in creating a meaningful association between the two modalities through shared semantic tokens.

To address the first task, in our model, visual features differentiate semantic items from the background spatially based on their associated sounds. This involves applying a multi-modal spatial attention between the source-aware audio embedding  $\mathbf{f}^{g}_{a_t}$ and the initial video encoding feature maps $\mathbf{X}^{r}_{P}$. By incorporating the source-aware video embedding $\mathbf{f}^{g}_{v_t}$, we derive the spatially-attended video representation $\mathbf{f}^{\mathrm{sa}}_{v_t}$:
\begin{align*} 
    & \mathbf{f}^{\mathrm{attn}}_{v_t} = \sigma({\mathbf{X}^{r}_{P}}_t^\intercal \circledast \mathbf{f}^{g}_{a_t})) \cdot {\mathbf{X}^{r}_{P}}_t, \\
    & \mathbf{f}^{\mathrm{sa}}_{v_t} = \mathrm{FC}(\mathrm{tanh}([\mathbf{f}^{g}_{v_t}; \mathbf{f}^{\mathrm{attn}}_{v_t}])),
\end{align*}

\noindent where $\circledast$ represents the convolution operation, which means this incorporating is broadcasting to all locations on the feature map. 

In practice, based on the computations above, we also observed the presence of contrastive information, allowing the model to better learn how to accurately extract semantic object embeddings spacially on the feature maps. Essentially, it is crucial not only allow the model to learn how to successfully align visual and audio information but also to penalize those errors in cases where visual and audio inputs do not belong to the same scene at all. This will ultimately enhance SaSR-Net's spatial attention capabilities.

To achieve this, during training, we supplement both a matched (positive) audio-video pair $\{(v_{t}, a_t)\}_{t=1}^{T}$  along with a mismatched (negative) pair,  $\{(v^{\prime}_{t}, a_t)\}_{t=1}^{T}$, where $v^{\prime}_{t}$ is from a 1-second random video clip that belongs to a different video  than $a_t$. 
Let $\mathbf{f}^{\mathrm{sa}}_{v_t}$ be the spatially-attended representation for a matched sample, and $\mathbf{f}^{\mathrm{sa}}_{v^{\prime}_t}$ be that for a mismatched sample. 
For the optimization of the traning process, we employ a loss function to distinguish between matched and mismatched samples using a binary classifier: 
\[
    \mathcal{L}_{\mathrm{match}}= \mathrm{CE}(\mathbf{f}^{\mathrm{sa}}_{v_t}, 1) + \mathrm{CE}(\mathbf{f}^{\mathrm{sa}}_{v_t}, 0).
\]
By minimizing this loss function, the learned representations become more discriminative.

\subsection{Multi-modal Temporal Attention}

\label{sec:multimodaltemporalattention}

In this section, we address the second task outlined in Sec.~\ref{sec:multimodalspecialattention}. 

Traditional QA methods treat questions as single entities, as in~\cite{alamri2019audio}. Our AVQA approach, however, utilizes the temporal sequences of data, such as frames and audio, to align questions with specific content moments. For example, a \textit{violin} query directs the focus to relevant video segments. This alignment leads to contextually accurate responses by linking question tokens to the correct temporal embeddings.

To achieve this, we introduce multi-modal temporal attention block that employs cross-attention through $t = 0$ to $T-1$ for updated audio embedding $\mathbf{f}^{\mathrm{ta}}_{A_T}$ and visual embedding  $\mathbf{f}^{\mathrm{ta}}_{V_T}$ based on the question's embedding $\mathbf{f}_{Q_L}$. The cross attention is calculated as follows,
\begin{align*}
    \mathbf{f}^{\mathrm{ta}}_{A_T} = \sigma(\frac{\mathbf{f}_{Q_L}{ \mathbf{f}^{g}_{A_T} }^\intercal}{\sqrt{D}})\mathbf{f}^{g}_{A_T},  \; \mathbf{f}^{g}_{A_T} = \{ \mathbf{f}^{g}_{a_t} \}_{t=1}^T ,
    \\\mathbf{f}^{\mathrm{ta}}_{V_T} = \sigma(\frac{\mathbf{f}_{Q_L}{\mathbf{f}^{\mathrm{sa}}_{V_T}}^\intercal}{\sqrt{D}})\mathbf{f}_{V_T}^{\mathrm{sa}}, \; \mathbf{f}^{\mathrm{sa}}_{V_T} = \{ \mathbf{f}^{\mathrm{sa}}_{v_t} \}_{t=1}^T. 
\end{align*}

\subsection{Answer Prediciton}
\label{sec:answerPrediction}

To predict the final answer to the question, we utilize the multi-modal temporal embeddings and semantically-enriched embeddings, as they have already been proven to contain competent high-dimensional values after attention masks. The implementation includes a shortcut connection structure and a necessary fusion network.

For the shortcut connection structure, we (averagely) reduce the semantically-enriched embeddings across their time dimension and aggregate them with the multi-modal temporal embeddings, modality by modality. This operation is expected to help maintain global information and facilitate gradient back-propagation.

We hope the fusion network could integrate both the audio-text modal and visual-text modal into a final mixed modal that could be directly taken advantage of by its classifier and output predictions. Hence, we concatenate the two embeddings after the shortcut connection structure and employ a fully-connected layer as a classifier to predict the answer. The full operation is formulated as follows,
\[
        \mathbf{f}_{av} = \mathrm{FC}(\mathrm{tanh}([ \mathbf{f}^{\mathrm{ta}}_{A_T} + \mathbf{f}^{g}_{A_T}; \mathbf{f}^{\mathrm{ta}}_{V_T} + \mathbf{f}^{g}_{V_T})]),
\]
\[
    \mathbf{\hat{y}} = \sigma(\mathrm{FC}(\mathrm{tanh}(\mathbf{f}_{av} \cdot \mathbf{f}_{Q_L}))).
\]
Here $\mathbf{y}$ denotes the right answer id encoded by an one-hot vector, and   $\mathbf{\hat{y}}$ represents the probabilities of selection among all the answers, to match $\mathbf{y}$ closely. Therefore, we use cross-entropy loss for AVQA to penalize incorrect predictions,
\[
\mathcal{L}_{\mathrm{avqa}} = \mathrm{CE}(\mathbf{y},\mathbf{\hat{y}}).
\]
At last, the overall training loss is:
\[
     \mathcal{L} = \mathcal{L}_{\mathrm{avqa}} + \lambda_1 \mathcal{L}_{\mathrm{source}} + \lambda_2 \mathcal{L}_{\mathrm{reg}} + \lambda_3 \mathcal{L}_{\mathrm{match}}. 
\]

\section{Experiment
\label{sec:pagestyle}}

\begin{table*}[h]
\begin{center}
\renewcommand{\arraystretch}{0.95}
\scalebox{0.7}{
\begin{tabular}{c|c|ccc|ccc|cccccc|c}
\hline
\multirow{2}{*}{Task}     & \multirow{2}{*}{Method} & \multicolumn{3}{c|}{Audio Question} & \multicolumn{3}{c|}{Visual Question} & \multicolumn{6}{c|}{Audio-Visual Question}                         & All   \\
                          &                         & Count   & Comp   & Avg.   & Count  & Local  & Avg.    & Exist & Local & Count & Comp & Temp & Avg.  & Avg.  \\ \hline
\multirow{2}{*}{AudioQA}  & FCNLSTM               & 70.45      & 66.22         & 68.88  & 63.89      & 46.74         & 55.21   & 82.01     & 46.28    & 59.34    & 62.15       & 47.33    & 60.06 & 60.34 \\
                          & CONVLSTM   & 73.55  & 67.17 & 71.20  & 67.17  & 55.84  & 61.44  & \underline{82.49} & 63.08  & 51.85  & 62.13  & 50.36  & 62.56  & 63.79  \\ \hline
\multirow{2}{*}{VisualQA} & HCAttn                 & 70.25      & 54.91         & 64.57  & 64.05      & 66.37         & 65.22   & 79.10       & 49.51    & 59.97    & 55.25       & 56.43    & 60.19 & 62.30 \\
                          & MCAN                   & \text{77.50}      & 55.24         & 69.25  & 71.56      & 70.93         & 71.24   & 80.40       & 54.48    & \text{64.91}    & 57.22       & 47.57    & 61.58 & 65.49 \\ \hline
\multirow{2}{*}{VideoQA}  & HME                & 74.76      & 63.56         & 70.61  & 67.97      & 69.46         & 68.76   & 80.30       & 53.18    & 63.19    & 62..69      & 59.83    & 64.05 & 66.45 \\
                          & HCRN             & 68.59      & 50.92         & 62.05  & 64.39      & 61.81         & 63.08   & 54.47       & 41.53    & 53.38    & 52.11       & 47.69    & 50.26 & 55.73 \\ \hline
\multirow{6}{*}{AVQA}     & AVSD~           &     72.41     &      61.90      &        68.52       &   67.39     &      74.19      &    70.83           &   81.61      &    58.79         &  63.89        &    61.52                 &    61.41      &    65.49   &   67.44    \\
                          & Pano-AVQA               & 74.36      & 64.56         & 70.73  & \text{69.39}      & \text{75.65}         & \text{72.56}   & 81.21       &  59.33   & \text{64.91}    & \text{64.22}       & 63.23    & \text{66.64} & \text{68.93} \\ 
                          & AVST  & 77.78  & 67.17 & \underline{73.87} & 73.52  & 75.27  & 74.40 & \underline{82.49} & 69.88  & 64.24  & 64.67  & 65.82  & 69.53  & 71.59 \\ 
                          &PSTP-Net  & 73.97  &  65.59  & 70.91& \textbf{77.15}  & \underline{77.36}  & \textbf{77.26}  & 76.18  & \underline{73.23}  & \underline{71.80}  & \underline{71.79}  & \underline{69.00}  & \underline{72.57}  & \underline{73.52}  \\
& TJSTG  & \textbf{80.38}  & 
\textbf{69.87}  &  \textbf{76.47} &  \underline{76.19}  & \textbf{77.55}  &  \underline{76.88}  & \textbf{82.59}  &  71.54  &  64.24  &  66.21  & 64.84  & 70.13  &  73.04  \\
\cline{2-15} 
                        &\textbf{SaSR-Net(ours)}  &  73.95  &\underline{69.81}& 73.56 & 73.76  & 71.84& 73.28 & 69.76  & \textbf{ 73.43}  & \textbf{73.64} & \textbf{79.15} & \textbf{77.46} & \textbf{74.66} & \textbf{74.21} \\ \hline
\end{tabular}
}
\caption{Different methods on Music-AVQA dataset. The top-2 results are highlighted.}
\label{cmp}
\end{center}
\end{table*}

\subsection{Experiments Setting}
\label{sec:format}

 \textbf{Datasets:} The MUSIC-AVQA dataset \cite{li2022learning} includes 9,290 videos, featuring 7,423 real and 1,867 synthetic examples, and 45,867 question-answer pairs. This dataset spans 9 audio-visual question types and 33 templates, showcasing 22 instruments categorized into Strings, Winds, Percussion, and Keyboards. Each video is annotated with instrument category labels. The dataset, designed for answering questions about the appearance, sounds, and associations of different objects in videos, is published under the Creative Commons Attribution-NonCommercial 4.0 International License and is public for research use. The question type primarily involves estimating answers.

The AVQA-Yang dataset~\cite{yang2022avqa} contains 57,015 videos paired with 57,335 questions that require understanding both audio and visual clues. The question type in this dataset is multiple-choice.

\noindent
\textbf{Implementation:} 
The audio data has a sampling rate of 16 $Hz$, and video data has 1 $fps$. Videos are segmented into non-overlapping 1-frame segments, each yielding a $512D$ feature vector. We sample 1-second video segments every 6 seconds. Audio segments, also 1-second long, are processed using a linear layer, converting them from $128D$ VGGish features to $512D$ feature vectors. Word embeddings are set to 512 dimensions. Our batch size is 16, and we train for 80 epochs using the Adam optimizer with an initial learning rate of $1e-4$, which decreases by a factor of 0.3 every 16 epochs. Also, we set $\lambda_1 = \lambda_2 =\lambda_3 = 0.5$.
Our model and related utility codes are based on PyTorch. We use \textit{torchinfo} to summary our model’s configuration. Our model contains 65,117,283 parameters (approximately 205.24 MB storage). We put our model trained as well as evaluated on an NVIDIA GeForce GTX 1080 Ti.

\noindent
\textbf{Evaluation:} Following \cite{li2022learning}, we use answer prediction accuracy as our evaluation metric.

\subsection{Comparison to Prior Work}


In this study, we introduced SaSR-Net, a novel multi-modal AVQA framework, and compared it with established unimodal and cross-modal question answering systems in Tab.~\ref{cmp} to demonstrate its effectiveness. The baselines include:
(1) Audio Question Answering: FCNLSTM~\cite{fayek2020temporal}, CONVLSTM~\cite{fayek2020temporal}.
(2) Visual Question Answering: HCAttn~\cite{lu2016hierarchical}, MCAN~\cite{yu2019deep}
(3) Video Question Answering: PSAC~\cite{li2019beyond}, HME~\cite{fan2019heterogeneous}, HCRN~\cite{le2020hierarchical}.
(4) Audio-Visual Question Answering: AVSD~\cite{schwartz2019simple}, Pano-AVQA~\cite{yun2021pano}, AVST~\cite{li2022learning}.
PSTP-Net~\cite{li2023progressive}
and TJSTG~\cite{jiang2023target}. 

These baselines primarily use general encoders to extract video features, which are then processed through attention mechanisms for question answering. In contrast, our SaSR-Net uses Source-wise Learnable Tokens to extract semantically compact features from videos and employs Source-aware Semantic Representation to align these with visual and audio features. This enhances the model's capability to integrate and understand individual sound sources and visual objects in AVQA queries, enriching the features semantically.

SaSR-Net not only delivers robust performance in audio and visual QA but also showcases exceptional results in audio-visual QA, a domain where previous AVQA methods have been less effective. We have made substantial improvements in this area. SaSR-Net excels particularly in Audio-Visual Questions, significantly outperforming AVST~\cite{li2022learning} with notable improvements in \textbf{Counting (3.55\%)}, \textbf{Localization (9.4\%)}, \textbf{Comparative (14.48\%)}, and \textbf{Temporal (11.64\%) } questions. Moreover, our method surpasses AVSD by 9.22\%, Pano-AVQA by 7.9\%, AVST by 5.13\%, PSTP-Net by 2.09\%, and TJSTG by 4.53\% in average accuracy,indicating a strong advancement in AVQA. In Audio QA, SaSR-Net achieves an average accuracy of 73.56\%, exceeding specialized models like FCNLSTM and CONVLSTM.

These exceptional results provide strong evidence of the effectiveness of our proposed Source-wise Learnable Tokens and Source-aware Semantic Representation. By embedding audio and visual features with semantic context relevant to the queries, these innovations significantly enhance the representational capabilities of the framework. The effective use of Source-wise Learnable Tokensfacilitates a deeper integration of audio and visual modalities, allowing SaSR-Net to accurately identify and address complex multimodal interactions inherent in AVQA tasks.

\begin{figure*}[h!]
\centering
\includegraphics[width=1\textwidth]{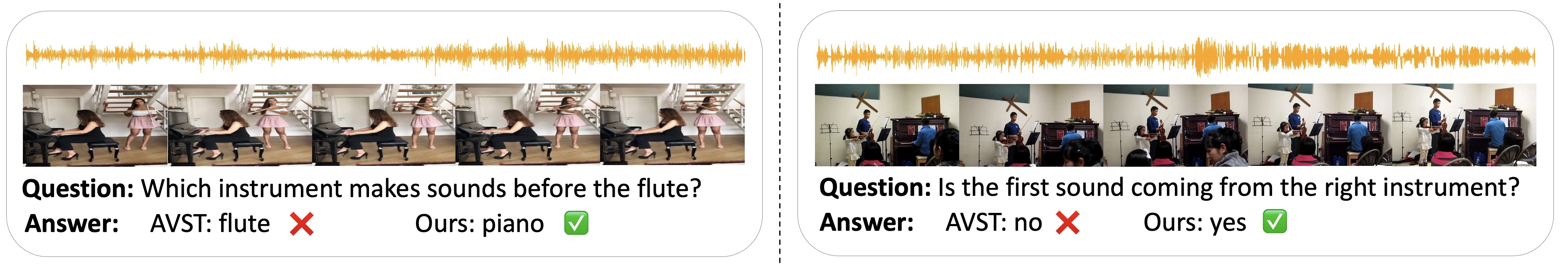}
\caption{Comparison of our SaSR-Net and AVST~\cite{li2022learning}. Our SaSR-Net provides more precise answers to complex questions by effectively integrating semantic information into audio and visual features.}

\label{case study}
\end{figure*}

\hfill

\subsection{Computational Efficiency}


In this section, we conducted performance comparisons on the Music-AVQA dataset to evaluate the computational efficiency of our SaSR-Net model in comparison with recent state-of-the-art AVQA methods: AVST~\cite{li2022learning}, PSTP-Net~\cite{li2023progressive} and TJSTG~\cite{jiang2023target}. 
Table~\ref{tab:method_comparison} summarizes the FLOPs and accuracy of each model.

\begin{table}[ht]
    \centering
    \footnotesize  
    \renewcommand{\arraystretch}{1}  
    \scalebox{0.92}{
    \begin{tabular}{c|c|c}
        \toprule
        \textbf{Method} & \textbf{FLOPs (G)} & \textbf{Acc (\%)} \\
        \midrule
        AVST ~\cite{li2022learning}      & 3.19  & 71.59 \\
        PSTP-Net ~\cite{li2023progressive} & 1.22  & 73.52 \\
        TJSTG ~\cite{jiang2023target}    & 1.22  & 73.52 \\
        SaSR-Net (ours) & 2.11  & 74.21 \\
        \bottomrule
    \end{tabular}
    }
    \caption{Comparison of methods by FLOPs and accuracy}
    \label{tab:method_comparison}
\end{table}

Our SaSR-Net achieves the highest accuracy of 74.21 with a moderate computational cost of 2.11 GFLOPs, balancing efficiency and performance. Although its FLOPs are slightly higher than those of PSTP-Net and TJSTG, which are both at 1.22 GFLOPs, SaSR-Net significantly outperforms them in accuracy. Compared to AVST, which requires 3.19 GFLOPs for a lower accuracy of 71.59, our model is both more efficient and more accurate. 

These results suggest that SaSR-Net is suitable for real-world applications where both accuracy and computational efficiency are important. The model's ability to achieve high performance with moderate computational requirements makes it practical for deployment in scenarios with limited computational resources.

\subsection{Ablation Studies}
\label{sec}



\begin{table}
\centering
\footnotesize  
\renewcommand{\arraystretch}{1}  
\scalebox{1}{
\begin{tabular}{cc|c|c}
\toprule
SLT & SaSR & Accuracy & Improvement \\
\midrule
\xmark & \xmark & 70.31\% & - \\
\xmark & \cmark & 71.78\% & $\uparrow$ 1.47\% \\
\cmark & \xmark & 72.16\% & $\uparrow$ 1.85\% \\
\cmark & \cmark & 74.21\% & $\uparrow$ 3.90\% \\
\bottomrule
\end{tabular}
}
\caption{Ablation on Source-wise Learnable Tokens (SLT) and Source-aware Semantic Representation (SaSR)}
\label{ablation1}
\end{table}

\begin{table}
\centering
\footnotesize  
\renewcommand{\arraystretch}{1}  
\scalebox{1}{
\begin{tabular}{cc|c|c}
\toprule
TA & SA & Accuracy & Improvement \\
\midrule
\xmark & \xmark & 70.17\% & - \\
\cmark & \xmark & 72.03\% & $\uparrow$ 1.03\% \\
\cmark & \cmark & 74.21\% & $\uparrow$ 2.18\% \\
\bottomrule
\end{tabular}
}
\caption{Ablation studies on Multi-modal Special Attention (SA), Multi-modal Temporal Attention (TA) blocks}
\label{ablation2}
\end{table}

In this section, we conducted ablation studies on Music-AVQA dataset to quantitatively evaluate the Source-wise Learnable Tokens (SLT) and the Source-aware Semantic Representation (SaSR) block, as presented in Table \ref{ablation1}. Additionally, we performed ablation studies to quantitatively assess the Multi-modal Spacial Attention (SA) and Multi-modal Temporal Attention (TA) blocks, as presented in Table \ref{ablation2}.

\noindent
\textbf{Effectiveness of SLT and SaSR:}
The inclusion and removal of the SLT (Source-wise Learnable Tokens ) and SaSR (Source-aware Semantic Representation) blocks impact the performance of the AVQA model. Removing both blocks leads to a considerable accuracy drop to 70.31\%. This decline occurs primarily because the model struggles to extract distinct semantic visual and auditory features without the SLT and fails to integrate these features without the SaSR, highlighting the critical roles these components play in comprehending complex audio-visual content. Conversely, introducing the SLT block in the baseline model increases the AVQA accuracy by 1.85\%, demonstrating its effectiveness in enhancing video comprehension by extracting more semantic information from diverse sources. Additionally, retaining the SaSR block while eliminating the SLT block results in a 1.47\% increase in accuracy, emphasizing the SaSR's crucial role in integrating diverse audio and visual features. More importantly, incorporating both SLT and SaSR into the model leads to a substantial improvement in accuracy by 3.90\%. These findings underscore the importance of both SLT and SaSR in aligning auditory elements with their corresponding visual cues and enhancing the model's question-answering capabilities.

\noindent
\textbf{Effectiveness of SA and TA:}
Removing the TA (Multi-modal Temporal Attention) and SA (Multi-modal Spatial Attention) blocks significantly reduces accuracy to 70.17\%, underscoring their importance. Without SA, the model cannot accurately locate sounding instruments in videos, and without TA, it struggles to understand temporal dynamics, severely impairing its ability to identify key frames and localize sound sources.
Introducing SA enhances the model’s ability to link sounding objects with their sounds in complex scenes, improving spatial precision. Adding TA helps align temporal sequences, pinpointing key video frames relevant to the query. Together, SA and TA increase AVQA accuracy by 1.03\%, highlighting their synergistic effect in boosting the model's comprehension of audio-visual content.

\begin{table}
\centering
\footnotesize  
\renewcommand{\arraystretch}{1}  
\scalebox{0.9}{
\begin{tabular}{c|c}
\hline
Method & Avg(\%) \\
\hline
HME~\cite{fan2019heterogeneous}+HAVF~\cite{yang2022avqa} & 85.0 \\ 
PSAC~\cite{li2019beyond}+HAVF~\cite{yang2022avqa} & 87.4 \\ 
LADNet~\cite{li2019learnable}+HAVF~\cite{yang2022avqa} & 84.1 \\ 
HGA~\cite{jiang2020reasoning}+HAVF~\cite{yang2022avqa} & 87.7 \\ 
HCRN~\cite{le2020hierarchical}+HAVF~\cite{yang2022avqa}& 89.0 \\ 
\hline
\textbf{SaSR-Net(ours)} & \textbf{89.9} \\
\hline
\end{tabular}
}
\caption{Results of different methods on AVQA-Yang dataset.}

\label{avqadataset}
\end{table}

\subsection{Visualization}
\label{sec:visualization}

\noindent
\textbf{Visualization of SA and TA:} In Fig.~\ref{sata}, we visualize the results of the Spatial Attention and Temporal Attention Blocks.

\noindent
\textbf{Comparative Results:}
In Fig.~\ref{case study}, we present the results of our SaSR-Net method, compared with the results of AVST~\cite{li2022learning}. Our approach more accurately answers complex questions with specific semantic information due to our SLT and SaSR blocks. The SLT extracts and aggregates semantic category information from various sources, while the SaSR effectively integrates these semantic-aware features into both audio and visual features. These aggregated featuires outperform the original features, leading to superior performance.

Previous AVQA methods often fail to accurately associate visual objects with corresponding sounds in complex scenes, leading to incorrect answers. In contrast, our SaSR-Net, with its SLT and SaSR blocks, effectively connects sounding objects with mixed audio sources and accurately pinpoints their locations using spatial attention. It also employs temporal attention to identify key timestamps related to the posed question. This enhances the model’s ability to map sound sources accurately, significantly improving audio-visual analysis in dynamic multi-modal environments.

\subsection{Experiments on AVQA Dataset}
\label{sec:avqa}
While most existing methods are tested on the MUSIC-AVQA dataset~\cite{li2022learning}, we extend the validation of our method to the AVQA-Yang dataset~\cite{yang2022avqa} to further demonstrate its effectiveness.  This confirms its applicability across different question formats and more complex scenarios. Following the approach in \cite{yang2022avqa}, we integrate various strategies \cite{fan2019heterogeneous, li2019beyond, li2019learnable, jiang2020reasoning, le2020hierarchical} with HAVF~\cite{yang2022avqa} as our evaluation metric. The comparative results in Table \ref{avqadataset} show that our method outperforms others on the AVQA dataset. This underscores the robustness of our proposed SaSR-Net in diverse audio-visual question answering environments.

\section{Conclusion}
\label{sec:foot}
In this paper, we present SaSR-Net, a novel AVQA approach that introduces source-aware learnable tokens to effectively capture and integrate semantic-aware audio-visual representations. This enhances alignment between audio elements and visual cues, crucial for identifying relevant scene regions and their association with questions. By excelling at extracting and understanding single-source information within complex scenes, SaSR-Net significantly improves performance on AVQA tasks.

\noindent
\textbf{Limitation:} While SaSR-Net achieves remarkable performance on multi-modal tasks, its results on single-modality (audio-only or visual-only) questions are not as outstanding. This may be due to training data bias, as the dataset contains a higher proportion of audio-visual questions, leading the model to be better tuned for multi-modal scenarios. To address this issue, we can fine-tune SaSR-Net on single-modality tasks, aiming to enhance its performance on audio-only and visual-only questions while maintaining its strong capabilities in multi-modal contexts.

Additionally, SaSR-Net may still face challenges in handling extremely noisy audio-visual data or scenarios with highly complex and overlapping audio sources. These situations could affect the model's ability to accurately extract and align semantic representations, highlighting areas for future improvement and research.



\bibliography{custom}

\appendix

\end{document}